\pdfoutput=1
\pdftrailerid{}
\pdfglyphtounicode{parenleftbig}{0028}\pdfglyphtounicode{parenrightbig}{0029}
\pdfglyphtounicode{parenleftBig}{0028}\pdfglyphtounicode{parenrightBig}{0029}
\pdfglyphtounicode{parenleftbigg}{0028}\pdfglyphtounicode{parenrightbigg}{0029}
\pdfglyphtounicode{parenleftBigg}{0028}\pdfglyphtounicode{parenrightBigg}{0029}
\pdfglyphtounicode{bracketleftbig}{005B}\pdfglyphtounicode{bracketrightbig}{005D}
\pdfglyphtounicode{bracketleftBig}{005B}\pdfglyphtounicode{bracketrightBig}{005D}
\pdfglyphtounicode{bracketleftbigg}{005B}\pdfglyphtounicode{bracketrightbigg}{005D}
\pdfglyphtounicode{bracketleftBigg}{005B}\pdfglyphtounicode{bracketrightBigg}{005D}
\pdfglyphtounicode{braceleftbig}{007B}\pdfglyphtounicode{bracerightbig}{007D}
\pdfglyphtounicode{braceleftBig}{007B}\pdfglyphtounicode{bracerightBig}{007D}
\pdfglyphtounicode{braceleftbigg}{007B}\pdfglyphtounicode{bracerightbigg}{007D}
\pdfglyphtounicode{braceleftBigg}{007B}\pdfglyphtounicode{bracerightBigg}{007D}
\pdfglyphtounicode{vextendsingle}{007C}\pdfglyphtounicode{vextenddouble}{2016}
\pdfglyphtounicode{summationtext}{2211}\pdfglyphtounicode{summationdisplay}{2211}
\pdfglyphtounicode{producttext}{220F}\pdfglyphtounicode{productdisplay}{220F}
\pdfglyphtounicode{integraltext}{222B}\pdfglyphtounicode{integraldisplay}{222B}
\documentclass[10pt,letterpaper]{article}

\usepackage{scenewright-arxiv}%

\usepackage{currfile} %

\usepackage{caption} %

\usepackage{multirow}
\usepackage{makecell}
\usepackage{colortbl}
\definecolor{groupband}{gray}{0.92}
\definecolor{heat}{HTML}{3B7FB3}   %

\hypersetup{
  pdfcreator={}, pdfproducer={},
  pdftitle={SceneWright: Benchmarking Coding Agents on Multi-View Procedural 3D Modeling of Outdoor Scenes},
  pdfauthor={Yuning Zhang, Haotian Chen, Yangmiao Zhang, Tiexin Li, Yuanbin Zhu, Xiaosong Gao, Binxuan Hu,
    Linxi Yu, Weikai Huang, Xinke Li},
}

\IfFileExists{cover/cover.tex}{\input{cover/cover.tex}}{}
\makeatletter
\@ifundefined{makecover}{%
  \PackageWarningNoLine{main}{cover/cover.tex not found or without \string\makecover;
    using the fallback title page}%
  \let\sw@fb@title\@empty \let\sw@fb@subtitle\@empty
  \let\sw@fb@authors\@empty \let\sw@fb@affils\@empty \let\sw@fb@links\@empty
  \def\swtitle#1{\gdef\sw@fb@title{#1}}%
  \def\swsubtitle#1{\gdef\sw@fb@subtitle{#1}}%
  \def\swauthor{\@ifnextchar[\sw@fb@author{\sw@fb@author[]}}%
  \def\sw@fb@author[#1]#2{%
    \ifx\sw@fb@authors\@empty\else\g@addto@macro\sw@fb@authors{,\quad}\fi
    \g@addto@macro\sw@fb@authors{#2\textsuperscript{#1}}}%
  \def\swaffiliation{\@ifnextchar[\sw@fb@affil{\sw@fb@affil[]}}%
  \def\sw@fb@affil[#1]#2#3{%
    \ifx\sw@fb@affils\@empty\else\g@addto@macro\sw@fb@affils{\quad}\fi
    \g@addto@macro\sw@fb@affils{\textsuperscript{#1}#2}}%
  \def\swlink#1#2{%
    \ifx\sw@fb@links\@empty\else\g@addto@macro\sw@fb@links{\quad$\cdot$\quad}\fi
    \g@addto@macro\sw@fb@links{\sw@fb@link{#1}{#2}}}%
  \def\sw@fb@link#1#2{%
    \def\sw@fb@name{#1}%
    \ifnum\pdfstrcmp{#1}{homepage}=0 \def\sw@fb@name{Homepage}\fi
    \ifnum\pdfstrcmp{#1}{huggingface}=0 \def\sw@fb@name{Hugging Face}\fi
    \ifnum\pdfstrcmp{#1}{github}=0 \def\sw@fb@name{GitHub}\fi
    \if\relax\detokenize{#2}\relax
      {\color{swink3}\sw@fb@name}%
    \else
      \href{#2}{\sw@fb@name}%
    \fi}%
  \def\makecover{%
    \thispagestyle{empty}%
    \begingroup
      \setlength{\parskip}{0pt}\centering
      \vspace*{36pt}%
      {\sw@sffont{24}{28}\sffamily\bfseries\color{swaccent}\sw@fb@title\par}%
      \vspace{8pt}%
      {\sw@sffont{15}{19}\sffamily\bfseries\color{swink}\sw@fb@subtitle\par}%
      \vspace{20pt}%
      {\sw@sffont{12}{15}\sffamily\color{swink}\sw@fb@authors\par}%
      \vspace{6pt}%
      {\sw@sffont{10}{13}\sffamily\color{swink2}\sw@fb@affils\par}%
      \vspace{26pt}%
      {\sw@sffont{12}{14}\sffamily\bfseries\color{swaccent}Abstract\par}%
      \vspace{8pt}%
      \parbox{0.86\textwidth}{Can a coding agent reconstruct an outdoor scene in code from calibrated views? We study \textbf{multi-view procedural 3D modeling}, in which an agent writes Blender code to build class-labeled entities in the metric frame of posed cameras. Existing evaluations fall short in this setting: scoring 2D rendered views can hide flawed 3D geometry, while aligning predicted meshes to ground truth leaves pose and scale untested. We introduce SceneWright, a benchmark of 50 script-built outdoor scenes paired with exact Blender ground truth. Each scene is posed as both a sparse-view image task and a dense survey-video task, for 100 tasks in total. Submitted scenes are scored directly against ground truth across four dimensions: layout, semantic composition, and geometric and visual fidelity. We test 11 models, each under its vendor CLI and under OpenCode, for 22 configurations. Our results show that the underlying model largely dictates the performance tier, while the effect of the agent harness varies from model to model, and higher token spend does not translate to better reconstructions. Agents differ far more in metric structure than in surface appearance, and under the current interface they gain little from a dense survey video over sparse views, using under 1\% of its frames. Finally, turning the inspection, iterative refinement, and photogrammetric practices of top-performing agents into harness skills lifts two weaker models without weight updates, pointing to the harness as a lever worth exploring.
}\par
      \vspace{16pt}%
      {\sw@sffont{9}{12}\sffamily\sw@fb@links\par}%
    \endgroup
    \clearpage}%
}{}
\makeatother
\providecommand*\swtitle[1]{}\providecommand*\swsubtitle[1]{}\providecommand*\swauthor[2][]{}
\providecommand*\swaffiliation[3][]{}\providecommand*\swlink[2]{}\providecommand*\swcornerlogo[1]{}

\swtitle{SceneWright}
\swsubtitle{Benchmarking Coding Agents on Multi-View Procedural 3D Modeling of Outdoor Scenes}
\usepackage{pifont}
\swauthor[1,5*]{Yuning Zhang}
\swauthor[1,5*]{Haotian Chen}
\swauthor[2,5*]{Yangmiao Zhang}
\swauthor[3]{Tiexin Li}
\swauthor[4]{Yuanbin Zhu}
\swauthor[4]{Xiaosong Gao}
\swauthor[4]{Binxuan Hu}
\swauthor[4]{Linxi Yu}
\swauthor[6]{Weikai Huang}
\swauthor[3\,\ding{41}]{Xinke Li}
\swaffiliation[1]{USTC}{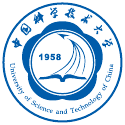}%
\swaffiliation[2]{UESTC}{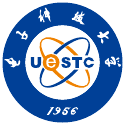}%
\swaffiliation[3]{CityUHK}{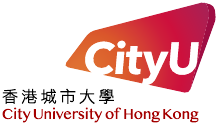}%
\swaffiliation[4]{SoTALab}{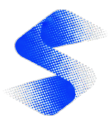}
\swaffiliation[5]{LLLab}{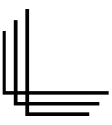}
\swaffiliation[6]{SJTU}{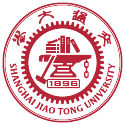}%
\swcornerlogo{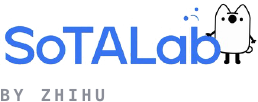}

\begin{document}

\makecover

\begin{figure}[t]
\centering
\includegraphics[width=0.90\linewidth]{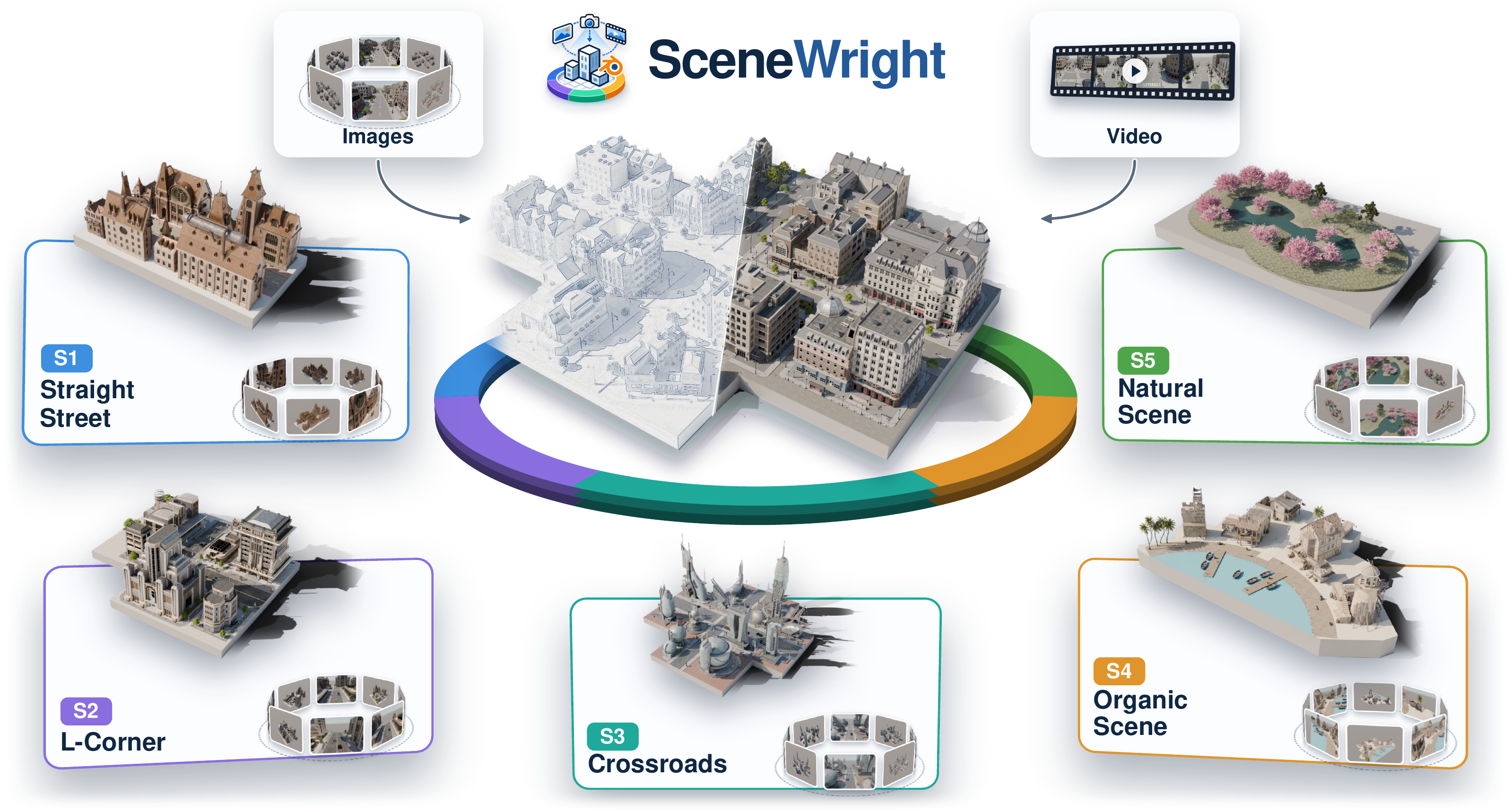}
\caption{\textbf{SceneWright.} A coding agent models an outdoor scene in Blender from calibrated observations, either selected images (T1, sparse views) or a survey video (T2, dense views), and submits it as an editable 3D model; the scene in the center is shown as plain geometry and with its materials. The 50 scenes are urban, in four layout types (S1--S4: straight street, L-shaped corner, crossroads, and organic scene), or natural (S5); each card shows an example scene with the ring of its input views.}
\label{fig:\currfilebase}%
\end{figure}

\makeatletter
\insert\footins{\reset@font\footnotesize
  \interlinepenalty\interfootnotelinepenalty \splittopskip\footnotesep \splitmaxdepth\dp\strutbox
  \floatingpenalty\@MM \hsize\columnwidth \@parboxrestore \parindent\z@
  \noindent\rule\z@\footnotesep\textsuperscript{*}Equal contribution.\par

  \noindent\textsuperscript{\ding{41}}Corresponding author.%
  \par
  \noindent This work was supported by SoTALab.\strut}%
\makeatother

\section{Introduction}
\label{sec:intro}

Building 3D models of outdoor environments, from city streets~\cite{dosovitskiy2017carla,somanath2024urban} to natural terrain~\cite{raistrick2023infinigen}, is important for interactive gaming and embodied simulation. These applications require scenes with both clear semantic structure and real-world physical scale, a combination long studied in procedural modeling~\cite{kelly2017bigsur,nishida2018procedural}. Recently, multimodal coding agents have begun creating 3D scenes by writing executable programs in graphics software~\cite{hu2024scenecraft,lu2025ll3m,yin2026viga}, taking text prompts and, in some systems, reference images as input~\cite{he2026seig,yang2026codeasroom,yin2026viga}. To test whether coding agents can faithfully model a specific observed 3D scene rather than generate a plausible one, we borrow the setting of multi-view reconstruction~\cite{schoenberger2016mvs,mildenhall2020nerf,kerbl20233dgs}, where posed cameras define a shared coordinate frame. The question is whether these agents can reason in 3D from calibrated views of an outdoor site and procedurally build semantically structured scenes aligned in a unified coordinate frame.

We define this problem as \emph{multi-view procedural 3D modeling} and focus on outdoor environments, especially urban street scenes. Existing agentic 3D benchmarks mainly target single objects or enclosed indoor rooms~\cite{gao2026threedcodebench,gu2025blendergym,liu2026threedharnessbench,yang2026p3dbench,zhao2026sceneactbench}; in contrast, outdoor and urban scenes involve much larger spatial extents, irregular road networks and terrain relief, and rich cross-scale semantic hierarchies, placing higher demands on global spatial reasoning. Many benchmarks also rely on brief prompts or single images~\cite{li2026lego,liu2025ir3dbench,lu2026worldcoderbench,ning2026vibeworlding,zheng2026voxelcodebench}, which leave much of the 3D scene unspecified and allow agents to synthesize any plausible layout. Multi-view capture instead imposes strict geometric constraints across views, requiring the agent to place roads or terrain, buildings, and objects at consistent positions, sizes, and orientations within a unified coordinate frame.

Benchmarks driven by text prompts or single images only need to check whether a synthesized scene looks plausible, while modeling a site from calibrated views requires verifying an exact 3D match against a unique ground truth. Existing protocols often score only rendered 2D views via image similarity or VLM judges~\cite{gao2026threedcodebench,gu2025blendergym,tang2026bvb}, which can miss geometric errors hidden behind plausible appearances~\cite{li2026lego,ye2026code4scene}. 3D evaluations that align predicted meshes to ground truth before scoring~\cite{liu2026threedharnessbench,zhao2026sceneactbench} leave pose and scale errors untested. A faithful evaluation must instead compare the submitted 3D scene directly against ground truth in the unified coordinate frame, and test appearance from held-out viewpoints. Such evaluation requires complete, unoccluded, and semantically labeled 3D ground truth. Because real-world outdoor scans rarely provide this, we construct a controlled synthetic environment where the ground truth is exact by construction.

We introduce \textbf{SceneWright}, a benchmark for multi-view procedural 3D modeling across 50 script-built outdoor scenes ranging up to 189\,m, each with exact Blender ground truth. Following standard multi-view capture regimes~\cite{long2022sparseneus,niemeyer2022regnerf,schoenberger2016mvs}, every scene is paired with sparse multi-view images (T1) and a dense survey video (T2), forming 100 tasks. Starting from an empty Blender file without external 3D assets, the agent writes Python code to build the scene, which we score in the unified coordinate frame without registration across layout ($E_1$), semantic composition ($E_2$), and geometric ($E_3$) and held-out visual ($E_4$) fidelity.

We evaluate 11 frontier models under their vendor CLIs and the OpenCode harness~\cite{opencode2026}. First, the model largely sets the performance tier, while the effect of changing the harness varies from model to model and token cost does not predict quality; even strong agents, though their renders look plausible, struggle with accurate spatial localization and geometry. Second, under the current interface, a dense survey video brings little gain over sparse views, and even the strongest agents use under 1\% of its frames. Third, trajectory analysis shows that the strongest agents rely on close view inspection, iterative render-and-compare refinement, and measurement from the given cameras; turning these practices into three harness skills improves two weaker models without updating model weights. Our contributions are:
\begin{itemize}
    \item \textbf{SceneWright}, a controlled benchmark for multi-view procedural 3D modeling across 50 urban and natural outdoor scenes under sparse and dense capture, with exact, class-labeled 3D ground truth in a unified coordinate frame;
    \item A registration-free, scene-level evaluation protocol spanning layout, semantic composition, and geometric and held-out visual fidelity, verified on reference scenes with known errors; and
    \item A study of 22 model-harness configurations that identifies bottlenecks in accurate spatial localization and dense multi-view utilization, and shows that harness skills drawn from strong trajectories can improve weaker agents without weight updates.
\end{itemize}

\section{Related Work}
\label{sec:related}
\begin{figure}[t]
\centering
\includegraphics[width=\textwidth]{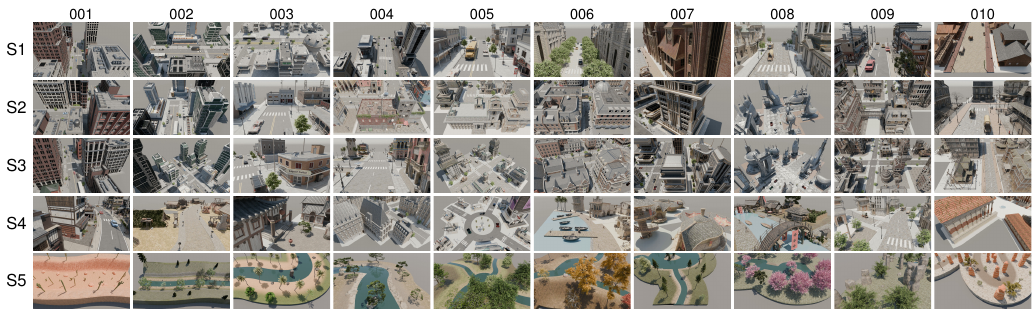}
\caption{\textbf{Overview of the 50 scenes.} Rows give the scene type (S1--S5) and columns the scene number.}
\label{fig:scenes50_vid}
\end{figure}

\runin{Structured and procedural modeling of outdoor scenes.}
Structured models of cities and landscapes have long been built by procedural rules or fitted to measurements. BigSUR~\cite{kelly2017bigsur} fuses street-level imagery, GIS footprints, and a coarse mesh in a global optimization and outputs city blocks as footprints, profiles, and facades with their windows, balconies, and roofs. Nishida \etal~\cite{nishida2018procedural} recover a procedural grammar from a single photograph of a building, from the building mass to its windows and doors. Urban digital twins generate building context procedurally from aerial imagery, footprints, and LiDAR~\cite{somanath2024urban}. These methods output labeled, editable scenes in metric units, as SceneWright expects, but rely on rules or optimizers designed for their domain. In SceneWright, a general-purpose coding agent must infer the structure and write the program itself, from calibrated views alone.

\runin{Multi-view reconstruction.}
Multi-view stereo recovers dense depth and normals from posed images~\cite{schoenberger2016mvs}, and neural radiance fields~\cite{mildenhall2020nerf} and 3D Gaussian splatting~\cite{kerbl20233dgs} optimize a scene representation so that its renders match the input images. With few views these methods lose geometry, which RegNeRF~\cite{niemeyer2022regnerf} counters by regularizing patches rendered from unobserved viewpoints and SparseNeuS~\cite{long2022sparseneus} by learned priors. Their evaluation compares surfaces within distance tolerances, as in the F-score of Tanks and Temples~\cite{knapitsch2017tanks}, and renders from held-out viewpoints. SceneWright adopts their calibrated cameras and both kinds of comparison, but the agent writes a program of labeled entities that cannot match the images pixel by pixel, so appearance is compared through features and color histograms rather than pixel-wise differences (\cref{sec:evaluation}).

\runin{Code-driven 3D modeling and its benchmarks.}
Agents increasingly create 3D content by programming graphics software~\cite{hu2024scenecraft,lu2025ll3m,wang2026adsl}, and some systems take reference images as input~\cite{yin2026viga,he2026seig,yang2026codeasroom}. These systems close the loop by comparing renders with the goal, and they succeed when the scene agrees with a prompt or a single image, which a plausible scene of roughly the right shape can achieve. SceneWright requires agreement with several calibrated views in metric 3D, so the agent must also measure; our skills (\cref{sec:harness}) make both render comparison and measurement systematic. Benchmarks cover single objects and scene edits~\cite{gu2025blendergym,gao2026threedcodebench,yang2026p3dbench,liu2026threedharnessbench}, scenes from single images~\cite{liu2025ir3dbench,li2026lego}, indoor scenes from uncalibrated videos~\cite{tang2026bvb}, dynamic scenes from videos~\cite{shen2026fourdcodebench}, game-engine scenes~\cite{ye2026code4scene}, worlds from language~\cite{ning2026vibeworlding,lu2026worldcoderbench,zheng2026voxelcodebench}, and rooms from calibrated views~\cite{zhao2026sceneactbench}. SceneWright targets a whole outdoor scene, from sparse views or a survey video, in the cameras' metric frame and without external 3D assets, the two observation regimes of sparse-view reconstruction~\cite{long2022sparseneus,niemeyer2022regnerf} and multi-view stereo~\cite{schoenberger2016mvs}.

\runin{Evaluating 3D outputs.}
Many benchmarks score renders by image similarity such as SSIM or LPIPS~\cite{wang2004ssim,zhang2018lpips}, VLM judges, or human preference~\cite{gu2025blendergym,gao2026threedcodebench,tang2026bvb,liu2026threedharnessbench}, which can hide structural errors~\cite{ye2026code4scene}; VLM judges pick the geometrically better of two renders at chance level~\cite{li2026lego}. Others inspect the scene~\cite{ye2026code4scene}, compare aligned geometry~\cite{liu2026threedharnessbench,zhao2026sceneactbench}, or score visible surfaces without registration~\cite{li2026lego}; SceneWright combines these, scoring the complete scene's visible surfaces without registration, per class, and appearance from held-out viewpoints (\cref{sec:evaluation}).

\runin{Agent harnesses.}
A model can perform differently under different harnesses, the code running its agent loop and tools~\cite{liu2026claudecode,lee2026metaharness}.
Benchmarks of 3D agents share one harness across models~\cite{zhao2026sceneactbench,tang2026bvb}, run each model in its vendor CLI~\cite{liu2026threedharnessbench}, or mix harnesses~\cite{shen2026fourdcodebench,ye2026code4scene}.
SceneActBench compared one model under two harnesses~\cite{zhao2026sceneactbench}; we compare 11 models under their vendor CLIs and OpenCode~\cite{opencode2026}, and use the harness to carry expert-inspired skills.

\section{SceneWright Benchmark}
\label{sec:benchmark}

SceneWright presents multi-view procedural 3D modeling on outdoor scenes (\cref{fig:overview}). We define the task (\cref{sec:formulation}) and describe the 50 scenes (\cref{fig:teaser,sec:scenes}), the sparse-view task T1 and the dense-view task T2 posed on each, with the quality control and coverage of the resulting 100 tasks (\cref{sec:tasks}). The submitted scene is compared directly with the ground truth using four complementary scores (\cref{sec:evaluation}), comprising layout ($E_1$), semantic composition ($E_2$), geometric fidelity ($E_3$), and visual fidelity ($E_4$).

\begin{figure*}[t]
\centering
\includegraphics[width=\linewidth]{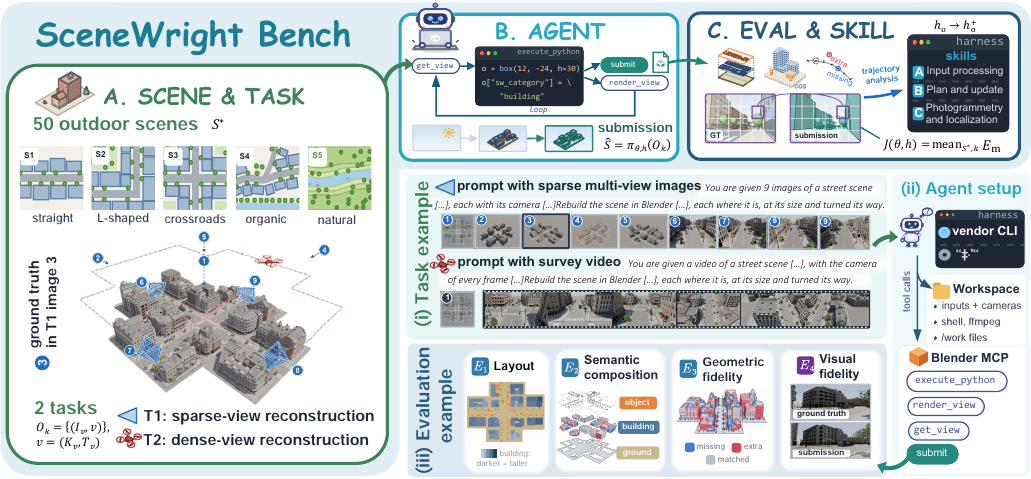}
\caption{\textbf{Overview of SceneWright.} (A)~Scenes and tasks: 50 outdoor scenes $S^\ast$, each posed as a sparse-view task (T1) and a dense-view task (T2) with calibrated observations $O_k$. (B)~The agent, a model $\theta$ under a harness $h$, writes and runs Blender code in a loop and submits a 3D model $\hat S=\pi_{\theta,h}(O_k)$. (C)~Evaluation and skills: $\hat S$ is compared with the ground truth by $E_1$--$E_4$ and summarized as $J(\theta,h)$; trajectory analysis distills three skills that the harness loads ($h_{\mathrm o}\to h_{\mathrm o}^{+}$). Below: (i)~task, (ii)~agent setup, and (iii)~evaluation examples.}
\label{fig:\currfilebase}
\end{figure*}

\subsection{Task Formulation}
\label{sec:formulation}

In multi-view procedural 3D modeling, each task gives the agent calibrated observations of a scene and the extent of its site, with all entities within the site visible across the observations (\cref{sec:qc}). The agent must model the exterior of the scene within the site, namely its ground and everything that stands on it. It builds this model as a Blender scene with its own code, in the metric frame of the cameras and without external 3D assets.

Formally, a scene is a set of entities in a metric frame, each a mesh with its materials and a class $y\in\mathcal{Y}$, with $\mathcal{Y}=\{\text{ground},\text{building},\text{object}\}$ in urban scenes and $\mathcal{Y}=\{\text{ground},\text{water},\text{object}\}$ in natural scenes; the ground truth $S^\ast$ is such a scene, given as a 3D model. Task $k\in\{\mathrm{T1},\mathrm{T2}\}$ provides the observations
\begin{equation}
O_k=\{(I_v,v):v\in\mathcal{C}_k\},\qquad v=(K_v,T_v),
\label{eq:observations}
\end{equation}
where each camera $v$ has known intrinsics $K_v$ and a pose $T_v\in SE(3)$ in the metric frame of $S^\ast$, and $I_v$ renders $S^\ast$ from $v$. The agent changes its scene only by executing code: starting from an empty Blender file $S_0$ that holds only the sun and sky of the inputs, its $t$-th program $c_t$, written from the observations and the replies of its tools, turns the scene into $S_t=c_t(S_{t-1})$. We evaluate a model $\theta$ together with the execution harness $h$ it runs in, which provides the model with its instructions and tools, and call $(\theta,h)$ a configuration. An episode of $(\theta,h)$ thus maps the observations to a submission,
\begin{equation}
\hat S=\pi_{\theta,h}(O_k),
\label{eq:episode}
\end{equation}
the agent's final submission, where $\pi_{\theta,h}$ stands for the entire episode.
Like $S^\ast$, $\hat S$ is a 3D model of separate, labeled entities that can be edited, measured in meters, and rendered from any camera. Therefore, it is scored as a scene,
\begin{equation}
s_k(\theta,h;S^\ast)=\bigl(E_1,E_2,E_3,E_4\bigr)(\hat S,S^\ast),
\label{eq:scores}
\end{equation}
where all four scores compare $\hat S$ and $S^\ast$ in the metric frame of $\mathcal{C}_k$ without registration, and two of them are computed from renders at evaluation cameras $\mathcal{C}^{\mathrm{ho}}$ (\cref{sec:evaluation}). Thus, pose and scale are part of what the agent must recover.

\begin{figure*}[p]
\centering
\includegraphics[width=\textwidth]{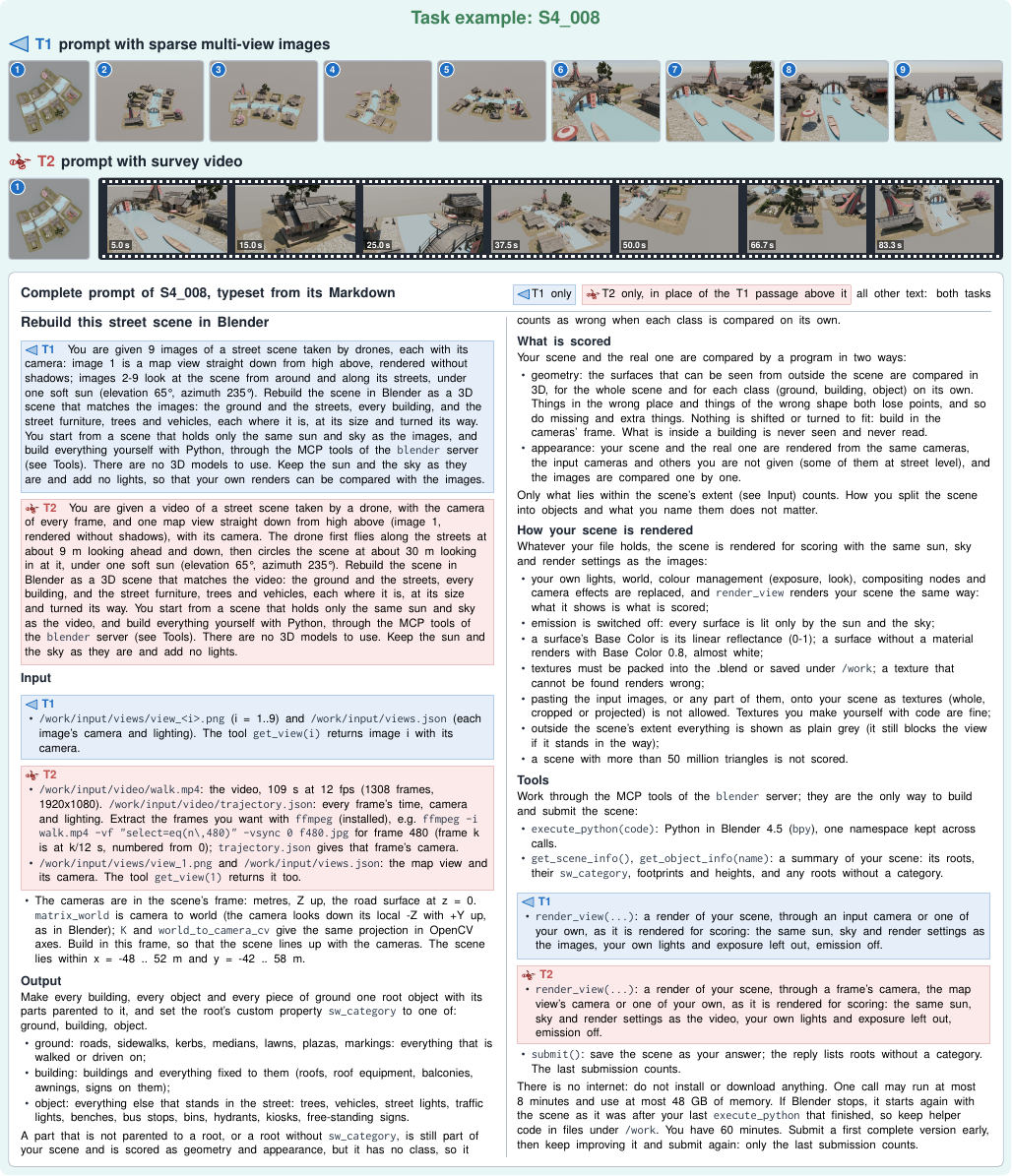}
\caption{\textbf{The two tasks of S4\_008 in full.} Top: the inputs, for T1 the nine images, each with its camera, and for T2 the same map view and the survey video, with the camera of every frame; seven frames are shown with their times. Bottom: both prompts, verbatim as the agent receives them under every harness, typeset from their Markdown; the T2 prompt equals the T1 prompt except for the passages on red, which replace the passages on blue before them.}
\label{fig:task_instance}
\end{figure*}

The agent works in Blender through tools served over MCP, among them a rendering tool that shows its current scene from any camera under the fixed lighting of the inputs and the rendering rules used for scoring, so that it can compare its work with the inputs. In T2, the agent extracts video frames with \texttt{ffmpeg}. Its tools and the rules of the task are given in \cref{fig:task_instance}.

\subsection{Scenes}
\label{sec:scenes}

\runin{Construction.} Every scene is assembled deterministically by its own script, with explicit positions and orientations, so its Blender ground truth is exact by construction. Across all scenes, structures of different classes and sizes share one metric frame, and every part is inferable from the inputs. For visual realism, buildings are taken from collected asset kits and placed by layout scripts, each scene drawing its buildings from one style, while natural terrain is built from real heightmaps or scripts and populated with assets from such kits.

\runin{Scene types.} This yields 50 outdoor scenes, comprising 359 buildings and 2{,}015 objects, with individual sites spanning up to 189\,m across. The 40 urban scenes fall into four layout types of ten scenes each, with increasing geometric complexity. They include a straight street (S1), lined by buildings on both sides; an L-shaped corner (S2), where two streets meet at a right angle around a corner building; and a crossroads (S3), where two streets cross, with a corner building in each quadrant. The organic scenes (S4) have curved streets, oblique junctions, squares, and waterfronts, and their buildings are not aligned to a common grid, so that the orientation of each building must be recovered as well as its position. The 10 natural scenes (S5) cover five landforms, with two scenes each, mirroring the urban types: straight valleys, L-shaped sea channels, and confluence pools for S1, S2, and S3, and lake islands and rock fields for S4.

\subsection{Sparse-View and Dense-View Tasks}
\label{sec:tasks}

Task T1 provides nine widely spaced, complementary views and Task T2 a survey video; in both, the agent models one complete scene in code. We treat these two tasks as distinct observation regimes, not as a controlled test of the number of views, since T1 and T2 also differ in their viewpoints and resolution. The instructions state what matters most and the rules of the task, but not the scoring formulas.

\runin{T1: sparse-view reconstruction.}
The nine images fall into three categories, captured by cameras placed via fixed rules. A top-down map view of the whole site (view~1) shows the overall layout of the scene. Four high oblique views from the site corners (views~2--5), each showing the whole site, capture the volumes and heights of buildings, trees, and terrain. Four low-altitude views (views~6--9) provide close-range details: urban scenes use street-level views, while natural scenes use low drone views along natural corridors.

\runin{T2: dense-view reconstruction.}
T2 provides the same top-down map view as T1 alongside a survey video from a continuous drone flight. The drone first flies at low altitude along the streets or natural corridors, and then circles above the entire site looking inward. \Cref{fig:scenes50_vid} shows one frame of every video. The agent receives this video paired with per-frame camera parameters.

\runin{Quality control.}
\label{sec:qc}%
Automated checks flag implausible layouts, which we then correct; the class, pose, and size of every building and object are also verified against the scene file. Since every task must be solvable and fair to all agents, cameras are placed to ensure that every entity is observed.

\subsection{Evaluating the Submitted Scene}
\label{sec:evaluation}

\label{sec:four_scores}
We compare the submitted scene directly with the ground truth using four complementary scores in $[0,1]$. $E_1$ and $E_2$ evaluate the visible 3D surface $\Gamma(S)$, whereas $E_3$ and $E_4$ evaluate renders from eight evaluation cameras $\mathcal{C}^{\mathrm{ho}}$, comprising four eye-level views along streets, paths, or shores and four oblique views from the sides of the site.

\label{sec:visible_surface}
Both scenes are voxelized and reduced to their visible surface $\Gamma(S)$, the exterior surfaces within the site seen from directions around and above it. $\Gamma_c(S)$ denotes the voxels of class $c\in\mathcal{K}$, namely ground, building, and object in urban scenes, with a channel class replacing building in natural scenes. Per-class quantities $x_c$ are combined across classes by their unweighted harmonic mean,
\begin{equation}
H_c(x_c)=\frac{|\mathcal{K}|}{\sum_{c\in\mathcal{K}}1/x_c},
\label{eq:hmean}
\end{equation}
so large ground or building surfaces do not mask missing objects. $H_c$ is zero if any $x_c$ is zero, as when no entity of the submission is labeled with class $c$.

\runin{Layout ($E_1$).}
$E_1$ evaluates the top-down layout and heights. We project $\Gamma(S)$ onto a plan grid, where each cell $q$ takes the class $m_S(q)$ of its highest visible voxel; $Q_S$ denotes the cells that hold visible building voxels of $S$, and $z_S(q)$ the top of these voxels in cell $q$. With hats and stars marking the submission and the ground truth, and $\mathrm{IoU}_c$ the intersection over union of the cells of class $c$ in $\hat m$ and $m^\ast$, $E_1$ averages the harmonic mean of the per-class 2D IoU on this grid and the mean height agreement $a_z$ over the union of building cells,
\begin{equation}
a_z=\frac{1}{|\hat Q\cup Q^\ast|}\sum_{q\in\hat Q\cap Q^\ast}\max\Bigl(0,1-\frac{|\hat z(q)-z^\ast(q)|}{\tau_z}\Bigr),\qquad
E_1=\tfrac12\bigl(H_c(\mathrm{IoU}_c)+a_z\bigr),
\label{eq:layout}
\end{equation}
with a height tolerance $\tau_z$. In natural scenes, $z_S(q)$ is the top of the ground or water in cell $q$, with a smaller $\tau_z$.

\runin{Semantic composition ($E_2$).}
Because the top-down plan omits vertical facades and object counts, $E_2$ averages three terms on the 3D visible surfaces $\Gamma_c(S)$, with $\hat P_c$ and $P^\ast_c$ the centers of their visible voxels of class $c$. The first is the harmonic mean of the per-class surface \mbox{F-score}~\cite{knapitsch2017tanks} over a set $\mathcal{T}$ of distance tolerances for local surface accuracy. With the precision $\mathrm{pr}_{c,\tau}$, the fraction of the points of $\hat P_c$ within $\tau$ of $P^\ast_c$, and the recall $\mathrm{re}_{c,\tau}$, the fraction of the points of $P^\ast_c$ within $\tau$ of $\hat P_c$, the F-score averages their harmonic mean over $\mathcal{T}$,
\begin{equation}
F_c=\frac{1}{|\mathcal{T}|}\sum_{\tau\in\mathcal{T}}\frac{2\,\mathrm{pr}_{c,\tau}\,\mathrm{re}_{c,\tau}}{\mathrm{pr}_{c,\tau}+\mathrm{re}_{c,\tau}}.
\label{eq:fscore}
\end{equation}
The second computes the harmonic mean of $\omega_c=\exp(-W_c/\lambda)$ from their per-class Wasserstein-1 distance $W_c$~\cite{rubner2000emd}, with a length scale $\lambda$, to capture coarse spatial shifts. The third is the F1 score $F_{\mathrm{obj}}$ of matching object instances in plan. Thus
\begin{equation}
E_2=\tfrac13\bigl(H_c(F_c)+H_c(\omega_c)+F_{\mathrm{obj}}\bigr).
\label{eq:composition}
\end{equation}

\runin{Geometric fidelity ($E_3$).}
To test visible geometry, both scenes are rendered from $\mathcal{C}^{\mathrm{ho}}$, which gives for every pixel ray $r$ the distance $d_S(r)$ to the first surface of $S$ along the ray and the normal $n_S(r)$ of that surface. Pooled over the eight cameras, $R_c$ denotes the rays whose hit in either scene is of class $c$; $G_c$ and $A_c$ are the rays of $R_c$ that the ground truth and the submission hit, $B_c=G_c\cap A_c$, and $M_c$ holds the rays of $B_c$ whose depths agree within $1\,\mathrm{m}$. Per class, depth agreement $\delta_c$ is the fraction of mutually hit rays whose first-hit depths agree within each tolerance of $\mathcal{T}$. Normal agreement $\nu_c$ sums the absolute dot product of surface normals on the rays of $M_c$, divides this sum once by $|G_c|$ and once by $|A_c|$, and averages the two ratios, so a ray hit by only one scene counts as zero:
\begin{equation}
\delta_c=\frac{1}{|\mathcal{T}|}\sum_{\tau\in\mathcal{T}}\frac{|\{r\in B_c:|\hat d(r)-d^\ast(r)|\le\tau\}|}{|B_c|},\qquad
\nu_c=\frac12\Bigl(\frac{1}{|G_c|}+\frac{1}{|A_c|}\Bigr)\sum_{r\in M_c}\bigl|\hat n(r)\cdot n^\ast(r)\bigr|.
\label{eq:geometry}
\end{equation}
$E_3$ averages the harmonic means of both terms, $E_3=\tfrac12\bigl(H_c(\delta_c)+H_c(\nu_c)\bigr)$.

\runin{Visual fidelity ($E_4$).}
To evaluate materials and appearance from held-out viewpoints, both scenes are rendered from $\mathcal{C}^{\mathrm{ho}}$ with their own materials under the inputs' lighting. $E_4$ averages the patchwise cosine similarity of DINOv2~\cite{oquab2024dinov2} features for structure and texture with the histogram IoU of chroma-weighted CIELAB $(a^\ast,b^\ast)$ distributions over foreground pixels for color. With $I(S;v)$ the scoring render of scene $S$ from camera $v$, $\phi_p$ the DINOv2 patch feature at grid position $p\in\mathcal{G}$, and $\eta_S(v)$ the histogram of the CIELAB chromaticity $(a^\ast,b^\ast)$ of the foreground pixels of either render,
\begin{equation}
\begin{gathered}
\Phi=\frac{1}{|\mathcal{C}^{\mathrm{ho}}|}\sum_{v\in\mathcal{C}^{\mathrm{ho}}}\frac{1}{|\mathcal{G}|}\sum_{p\in\mathcal{G}}\cos\bigl(\phi_p(I(\hat S;v)),\phi_p(I(S^\ast;v))\bigr),\\
\kappa=\frac{1}{|\mathcal{C}^{\mathrm{ho}}|}\sum_{v\in\mathcal{C}^{\mathrm{ho}}}\frac{\sum_b\min\bigl(\eta_{\hat S}(v)_b,\eta_{S^\ast}(v)_b\bigr)}{\sum_b\max\bigl(\eta_{\hat S}(v)_b,\eta_{S^\ast}(v)_b\bigr)},\qquad
E_4=\tfrac12(\Phi+\kappa).
\end{gathered}
\label{eq:visual}
\end{equation}

\runin{Aggregation.}
\label{sec:scores_show}
A submission is summarized by $E_{\mathrm{m}}$, the mean of $E_1$--$E_4$. A configuration's result on task $k$ averages each of $E_1$--$E_4$ and $E_{\mathrm{m}}$ over the 50 scenes $\mathcal{S}$ with equal weight. A bar marks the mean over T1 and T2 (all 100 tasks), as in $\bar{E}_1$, and the overall result, called Overall and written $J$, is $\bar{E}_{\mathrm{m}}$,
\begin{equation}
\begin{gathered}
E_{\mathrm{m}}=\tfrac14\bigl(E_1+E_2+E_3+E_4\bigr),\\
J(\theta,h)=\bar{E}_{\mathrm{m}}(\theta,h)=\frac{1}{2|\mathcal{S}|}\sum_{k}\sum_{S^\ast\in\mathcal{S}}E_{\mathrm{m}}(\hat S,S^\ast).
\end{gathered}
\label{eq:aggregate}
\end{equation}

\runin{Checking the scores.}
\label{sec:score_checks}
Reference scenes with known errors confirm that the scores do not depend on how a scene is split into objects and capture distinct failure modes. Gray materials lower only $E_4$, whereas shifting an urban scene by $1\,\mathrm{m}$ keeps $E_2$ and $E_4$ high but drops $E_3$.

\section{Expert-Inspired Skills}
\label{sec:harness}
A user can change the harness $h$ but not the model $\theta$. We therefore ask whether a weaker model can be improved through its harness. We take the two strongest models in our evaluation as experts and analyze 40 of their runs for how they view the inputs and measure the scene.

\subsection{Three Skills}
\label{sec:three_skills}
\begin{figure*}[t]
\centering
\includegraphics[width=\textwidth]{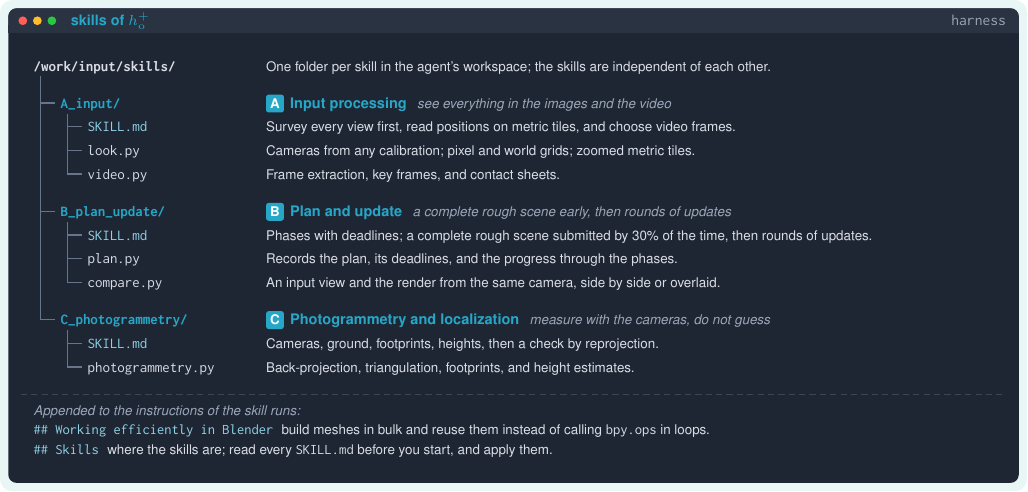}
\caption{\textbf{The three skills.} The skill folders that the agent finds in its workspace under OpenCode with the skills ($h_{\mathrm{o}}^{+}$, \cref{sec:skills_harness}), and what each file contains: each \texttt{SKILL.md} states a procedure, and the scripts carry out its steps. The instructions of the skill runs add the two paragraphs at the bottom.}
\label{fig:skills}
\end{figure*}

Each skill is a folder in the agent's workspace (\cref{fig:skills}).

\runin{Skill A: Input processing.}
A model sees large images downscaled and a long video only through a few frames, so small objects and many sides of the scene go unseen. Skill~A has the agent survey every view first, then read positions on full-resolution tiles with a metric grid and find each object in every view that shows it. In T2, it selects video frames by camera pose, target, or visual change and uses them as calibrated views.

\runin{Skill B: Plan and update.}
Agents often spend the budget on one part of the scene or stop once a first version exists. Skill~B has the agent submit a complete rough scene early and then revise it in rounds. In each round, the agent renders the scene from an input camera, compares the render with that input image, and re-measures the items that differ most. The agent keeps a change only if the render then matches better, and it continues until the budget is used.

\runin{Skill C: Photogrammetry and localization.}
Positions and heights estimated by eye can be off by meters. Skill~C has the agent compute them from the given cameras. The agent back-projects pixels at known heights, turns corners into footprints, and triangulates points across views. A measurement is accepted only after reprojection into another view, and every item is built from measured values rather than estimates.
\subsection{Loading Skills through the Harness}
\label{sec:skills_harness}
The skills are loaded through the harness: adding them to OpenCode ($h_{\mathrm{o}}$) turns it into $h_{\mathrm{o}}^{+}$, with the model unchanged.
They contain no ground truth, evaluator, scores, or benchmark statistics, and we test them in \cref{sec:skills_eval}.
\begin{figure*}[t]
\centering
\includegraphics[width=\linewidth]{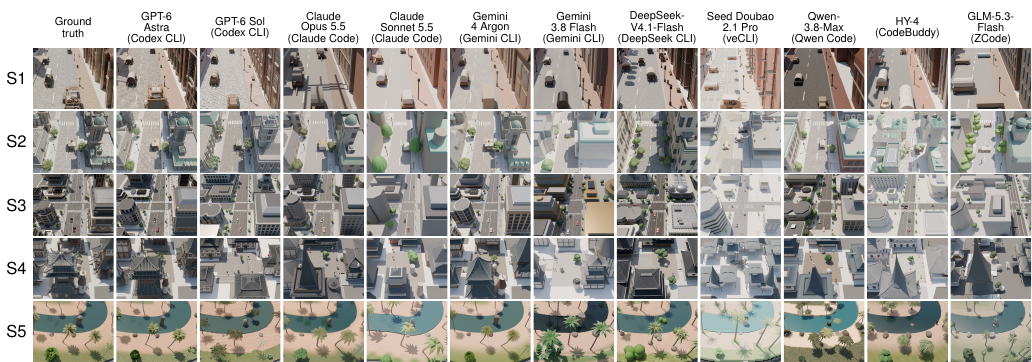}
\caption{\textbf{Submitted scenes.} The ground truth and the scenes submitted in T1 by the 11 models, each under its vendor CLI, on one scene of each type (S1--S5), rendered from the same viewpoint.}
\label{fig:\currfilebase}
\end{figure*}

\section{Experiments}
\label{sec:experiments}
\begin{table*}[t!]
\centering
\caption{\textbf{Main results on SceneWright.} Each model runs under its vendor CLI (upper block; \cref{sec:setup}) and under OpenCode (lower block). For each task, $E_1$--$E_4$ (\cref{sec:four_scores}) and their mean $E_{\mathrm{m}}$ are averaged over the 50 scenes; Overall averages $E_{\mathrm{m}}$ over the two tasks ($J=\bar{E}_{\mathrm{m}}$ in \cref{eq:aggregate}). Within each block, the best score in a column is in \textbf{bold} and the second best \underline{underlined}; cell shading grows with the score, scaled per column over all 22 rows. Tok.\ (input tokens, including cache reads, plus output tokens) and Time are totals over the 100 runs of both tasks. $\uparrow$/$\downarrow$: higher/lower is better.}
\label{tab:main}
\small
\setlength{\tabcolsep}{4.1pt}
\renewcommand{\arraystretch}{1.08}
\begin{tabular}{@{}l ccccc ccccc c rr@{}}
\toprule
\multirow{2}{*}{Model}
 & \multicolumn{5}{c}{T1 (sparse views)} & \multicolumn{5}{c}{T2 (dense views)} & \multirow{2}{*}{\makecell{Overall\\$J(\theta,h)\uparrow$}}
 & \multirow{2}{*}{\makecell{Tok.\\(M)$\downarrow$}} & \multirow{2}{*}{\makecell{Time\\(h)$\downarrow$}} \\
\cmidrule(lr){2-6} \cmidrule(lr){7-11}
 & $E_1\uparrow$ & $E_2\uparrow$ & $E_3\uparrow$ & $E_4\uparrow$ & $E_{\mathrm{m}}\uparrow$ & $E_1\uparrow$ & $E_2\uparrow$ & $E_3\uparrow$ & $E_4\uparrow$ & $E_{\mathrm{m}}\uparrow$ & & & \\
\midrule
\rowcolor{groupband}\multicolumn{14}{c}{\textbf{Vendor CLI}} \\
GPT-6 Astra & \cellcolor{heat!46}\textbf{0.605} & \cellcolor{heat!46}\textbf{0.714} & \cellcolor{heat!46}\textbf{0.524} & \cellcolor{heat!46}\textbf{0.751} & \cellcolor{heat!46}\textbf{0.648} & \cellcolor{heat!46}\textbf{0.640} & \cellcolor{heat!46}\textbf{0.715} & \cellcolor{heat!46}\textbf{0.551} & \cellcolor{heat!46}\textbf{0.732} & \cellcolor{heat!46}\textbf{0.660} & \cellcolor{heat!46}\textbf{0.654} & 777 & 85.84 \\
GPT-6 Sol & \cellcolor{heat!22}0.351 & \cellcolor{heat!24}0.496 & \cellcolor{heat!17}0.245 & \cellcolor{heat!24}0.618 & \cellcolor{heat!21}0.428 & \cellcolor{heat!20}0.364 & \cellcolor{heat!23}0.501 & \cellcolor{heat!17}0.252 & \cellcolor{heat!25}0.619 & \cellcolor{heat!21}0.434 & \cellcolor{heat!20}0.431 & 1190 & 68.14 \\
Claude Opus 5.5 & \cellcolor{heat!28}0.414 & \cellcolor{heat!29}0.539 & \cellcolor{heat!24}0.318 & \cellcolor{heat!25}0.624 & \cellcolor{heat!27}0.474 & \cellcolor{heat!31}0.485 & \cellcolor{heat!33}0.591 & \cellcolor{heat!28}0.368 & \cellcolor{heat!26}0.624 & \cellcolor{heat!30}0.517 & \cellcolor{heat!28}0.495 & 2660 & 68.78 \\
Claude Sonnet 5.5 & \cellcolor{heat!30}0.438 & \cellcolor{heat!33}0.583 & \cellcolor{heat!26}0.331 & \cellcolor{heat!18}0.581 & \cellcolor{heat!28}0.483 & \cellcolor{heat!26}0.428 & \cellcolor{heat!31}0.573 & \cellcolor{heat!23}0.319 & \cellcolor{heat!19}0.587 & \cellcolor{heat!25}0.477 & \cellcolor{heat!26}0.480 & 101 & 5.63 \\
Gemini 4 Argon & \cellcolor{heat!43}\underline{0.572} & \cellcolor{heat!43}\underline{0.686} & \cellcolor{heat!35}\underline{0.424} & \cellcolor{heat!35}\underline{0.687} & \cellcolor{heat!40}\underline{0.592} & \cellcolor{heat!41}\underline{0.588} & \cellcolor{heat!43}\underline{0.693} & \cellcolor{heat!36}\underline{0.452} & \cellcolor{heat!40}\underline{0.698} & \cellcolor{heat!40}\underline{0.608} & \cellcolor{heat!40}\underline{0.600} & 675 & 91.34 \\
Gemini 3.8 Flash & \cellcolor{heat!18}0.311 & \cellcolor{heat!21}0.463 & \cellcolor{heat!11}0.191 & \cellcolor{heat!13}0.551 & \cellcolor{heat!16}0.379 & \cellcolor{heat!17}0.339 & \cellcolor{heat!21}0.485 & \cellcolor{heat!13}0.213 & \cellcolor{heat!15}0.563 & \cellcolor{heat!17}0.400 & \cellcolor{heat!16}0.390 & 1352 & 47.04 \\
DeepSeek-V4.1-Flash & \cellcolor{heat!16}0.297 & \cellcolor{heat!16}0.413 & \cellcolor{heat!10}0.186 & \cellcolor{heat!19}0.585 & \cellcolor{heat!15}0.370 & \cellcolor{heat!15}0.312 & \cellcolor{heat!16}0.436 & \cellcolor{heat!12}0.203 & \cellcolor{heat!22}0.601 & \cellcolor{heat!15}0.388 & \cellcolor{heat!15}0.379 & 3315 & 45.99 \\
Seed Doubao 2.1 Pro & \cellcolor{heat!4}0.167 & \cellcolor{heat!4}0.290 & \cellcolor{heat!4}0.126 & \cellcolor{heat!5}0.506 & \cellcolor{heat!4}0.272 & \cellcolor{heat!6}0.216 & \cellcolor{heat!7}0.347 & \cellcolor{heat!6}0.146 & \cellcolor{heat!4}0.503 & \cellcolor{heat!6}0.303 & \cellcolor{heat!4}0.287 & 1042 & 73.64 \\
Qwen-3.8-Max & \cellcolor{heat!23}0.368 & \cellcolor{heat!27}0.524 & \cellcolor{heat!18}0.257 & \cellcolor{heat!20}0.592 & \cellcolor{heat!22}0.435 & \cellcolor{heat!19}0.359 & \cellcolor{heat!24}0.507 & \cellcolor{heat!17}0.251 & \cellcolor{heat!21}0.593 & \cellcolor{heat!20}0.428 & \cellcolor{heat!20}0.431 & 678 & 71.72 \\
HY-4 & \cellcolor{heat!17}0.301 & \cellcolor{heat!18}0.428 & \cellcolor{heat!9}0.176 & \cellcolor{heat!16}0.572 & \cellcolor{heat!15}0.369 & \cellcolor{heat!14}0.305 & \cellcolor{heat!15}0.429 & \cellcolor{heat!10}0.185 & \cellcolor{heat!16}0.570 & \cellcolor{heat!14}0.372 & \cellcolor{heat!14}0.371 & 1040 & 97.66 \\
GLM-5.3-Flash & \cellcolor{heat!12}0.252 & \cellcolor{heat!11}0.364 & \cellcolor{heat!6}0.141 & \cellcolor{heat!8}0.524 & \cellcolor{heat!9}0.320 & \cellcolor{heat!4}0.200 & \cellcolor{heat!4}0.322 & \cellcolor{heat!4}0.123 & \cellcolor{heat!4}0.504 & \cellcolor{heat!4}0.287 & \cellcolor{heat!6}0.304 & 974 & 85.91 \\
\rowcolor{groupband}\multicolumn{14}{c}{\textbf{OpenCode}} \\
GPT-6 Astra & \cellcolor{heat!44}\underline{0.580} & \cellcolor{heat!46}\textbf{0.716} & \cellcolor{heat!43}\textbf{0.494} & \cellcolor{heat!43}\textbf{0.730} & \cellcolor{heat!44}\textbf{0.630} & \cellcolor{heat!42}\textbf{0.600} & \cellcolor{heat!46}\textbf{0.718} & \cellcolor{heat!41}\textbf{0.505} & \cellcolor{heat!45}\textbf{0.724} & \cellcolor{heat!43}\textbf{0.637} & \cellcolor{heat!44}\textbf{0.633} & 941 & 85.71 \\
GPT-6 Sol & \cellcolor{heat!14}0.271 & \cellcolor{heat!20}0.452 & \cellcolor{heat!9}0.172 & \cellcolor{heat!20}0.592 & \cellcolor{heat!15}0.372 & \cellcolor{heat!14}0.310 & \cellcolor{heat!21}0.483 & \cellcolor{heat!11}0.198 & \cellcolor{heat!20}0.588 & \cellcolor{heat!16}0.395 & \cellcolor{heat!15}0.383 & 669 & 59.55 \\
Claude Opus 5.5 & \cellcolor{heat!27}0.408 & \cellcolor{heat!28}0.531 & \cellcolor{heat!24}0.318 & \cellcolor{heat!22}0.606 & \cellcolor{heat!26}0.466 & \cellcolor{heat!32}0.489 & \cellcolor{heat!35}0.610 & \cellcolor{heat!29}0.379 & \cellcolor{heat!31}0.652 & \cellcolor{heat!32}0.533 & \cellcolor{heat!28}0.499 & 1413 & 75.21 \\
Claude Sonnet 5.5 & \cellcolor{heat!21}0.349 & \cellcolor{heat!22}0.470 & \cellcolor{heat!18}0.260 & \cellcolor{heat!17}0.576 & \cellcolor{heat!20}0.414 & \cellcolor{heat!21}0.376 & \cellcolor{heat!23}0.497 & \cellcolor{heat!19}0.279 & \cellcolor{heat!18}0.581 & \cellcolor{heat!20}0.433 & \cellcolor{heat!20}0.424 & 212 & 11.78 \\
Gemini 4 Argon & \cellcolor{heat!46}\textbf{0.600} & \cellcolor{heat!45}\underline{0.710} & \cellcolor{heat!42}\underline{0.484} & \cellcolor{heat!38}\underline{0.702} & \cellcolor{heat!43}\underline{0.624} & \cellcolor{heat!40}\underline{0.572} & \cellcolor{heat!43}\underline{0.691} & \cellcolor{heat!36}\underline{0.449} & \cellcolor{heat!36}\underline{0.676} & \cellcolor{heat!39}\underline{0.597} & \cellcolor{heat!41}\underline{0.610} & 738 & 93.39 \\
Gemini 3.8 Flash & \cellcolor{heat!21}0.346 & \cellcolor{heat!23}0.483 & \cellcolor{heat!13}0.208 & \cellcolor{heat!15}0.563 & \cellcolor{heat!18}0.400 & \cellcolor{heat!18}0.346 & \cellcolor{heat!22}0.493 & \cellcolor{heat!14}0.221 & \cellcolor{heat!14}0.558 & \cellcolor{heat!17}0.404 & \cellcolor{heat!17}0.402 & 1442 & 51.81 \\
DeepSeek-V4.1-Flash & \cellcolor{heat!13}0.266 & \cellcolor{heat!15}0.400 & \cellcolor{heat!9}0.171 & \cellcolor{heat!15}0.565 & \cellcolor{heat!13}0.350 & \cellcolor{heat!11}0.274 & \cellcolor{heat!14}0.412 & \cellcolor{heat!9}0.174 & \cellcolor{heat!18}0.582 & \cellcolor{heat!12}0.360 & \cellcolor{heat!12}0.355 & 2080 & 47.63 \\
Seed Doubao 2.1 Pro & \cellcolor{heat!8}0.209 & \cellcolor{heat!10}0.355 & \cellcolor{heat!6}0.141 & \cellcolor{heat!4}0.497 & \cellcolor{heat!7}0.300 & \cellcolor{heat!7}0.233 & \cellcolor{heat!8}0.359 & \cellcolor{heat!6}0.148 & \cellcolor{heat!5}0.510 & \cellcolor{heat!7}0.312 & \cellcolor{heat!6}0.306 & 712 & 93.40 \\
Qwen-3.8-Max & \cellcolor{heat!21}0.344 & \cellcolor{heat!24}0.490 & \cellcolor{heat!15}0.229 & \cellcolor{heat!20}0.595 & \cellcolor{heat!20}0.414 & \cellcolor{heat!19}0.355 & \cellcolor{heat!22}0.496 & \cellcolor{heat!15}0.231 & \cellcolor{heat!19}0.584 & \cellcolor{heat!19}0.416 & \cellcolor{heat!19}0.415 & 721 & 69.31 \\
HY-4 & \cellcolor{heat!19}0.320 & \cellcolor{heat!19}0.447 & \cellcolor{heat!11}0.191 & \cellcolor{heat!18}0.584 & \cellcolor{heat!17}0.386 & \cellcolor{heat!17}0.333 & \cellcolor{heat!19}0.464 & \cellcolor{heat!13}0.214 & \cellcolor{heat!20}0.592 & \cellcolor{heat!17}0.401 & \cellcolor{heat!16}0.393 & 1459 & 93.15 \\
GLM-5.3-Flash & \cellcolor{heat!13}0.263 & \cellcolor{heat!14}0.396 & \cellcolor{heat!7}0.153 & \cellcolor{heat!10}0.534 & \cellcolor{heat!11}0.337 & \cellcolor{heat!7}0.231 & \cellcolor{heat!7}0.347 & \cellcolor{heat!5}0.137 & \cellcolor{heat!5}0.511 & \cellcolor{heat!6}0.306 & \cellcolor{heat!8}0.322 & 888 & 87.99 \\
\bottomrule
\end{tabular}
\end{table*}

\begin{figure*}[t]
\centering
\includegraphics[width=\linewidth]{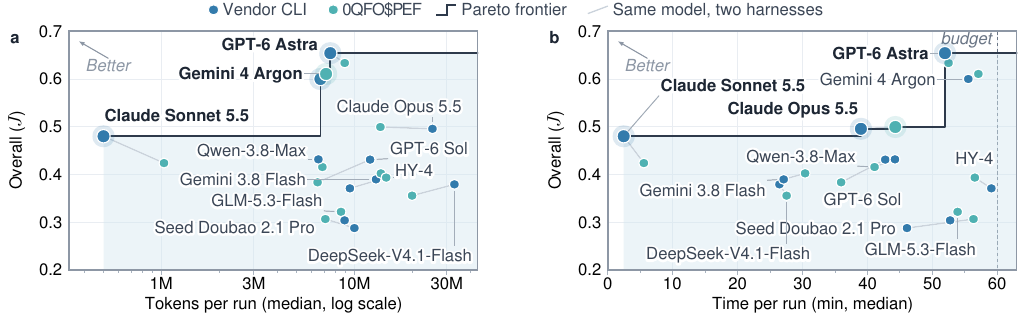}
\caption{\textbf{Score versus tokens and time.} Overall score ($J$ in \cref{eq:aggregate}) of the 22 configurations against (a)~the median tokens per run (input, including cache reads, and output) and (b)~the median time per run. Larger markers and bold labels mark the observed Pareto frontier.}
\label{fig:\currfilebase}
\end{figure*}
\begin{figure*}[t]
\centering
\includegraphics[width=\linewidth]{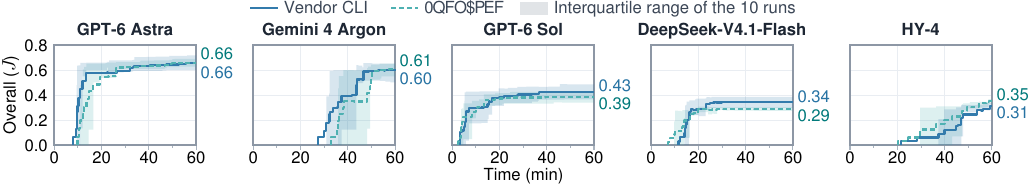}
\caption{\textbf{Score in hand over the hour.} At each moment of the 60-minute budget, every run is scored by its latest submission (0 before the first), and a line gives $J$ (\cref{eq:aggregate}) over 10 runs, the first scene of each type (S1--S5) in T1 and T2; it leaves the axis when the first of these runs submits. Bands are darker for the vendor CLI; the numbers at the right are the final values, which \cref{tab:main} reports over all 100 runs.}
\label{fig:trajectory_models}
\end{figure*}
We describe the setup (\cref{sec:setup}), compare models, harnesses, and tasks (\cref{sec:main_results}), relate scores to cost and time (\cref{sec:efficiency}), and test whether the practices of the strongest agents help weaker ones as skills (\cref{sec:skills_eval}).
\subsection{Setup}
\label{sec:setup}
\runin{Configurations.}
We evaluate 11 models, each under its vendor CLI and under OpenCode~\cite{opencode2026}, a general-purpose harness that runs every model through the same interface: 22 configurations, written harness / model. The vendor CLIs are Claude Code for Claude Opus 5.5 and Claude Sonnet 5.5, Codex CLI for GPT-6 Astra and GPT-6 Sol, Gemini CLI for Gemini 4 Argon and Gemini 3.8 Flash, DeepSeek CLI for DeepSeek-V4.1-Flash, CodeBuddy for HY-4, Qwen Code for Qwen-3.8-Max, ZCode for GLM-5.3-Flash, and veCLI for Seed Doubao 2.1 Pro.

\runin{Run settings.}
Both configurations of a model receive the same instructions, inputs, tools, and 60-minute budget and run in the same isolated environment, without human intervention. Every model runs at high reasoning effort. Each configuration runs all 100 tasks once, for 2{,}200 runs in total. Every run is scored on the scene it submitted; we report the four scores and their mean $E_{\mathrm{m}}$ over the 50 scenes (\cref{eq:aggregate}), with time and tokens.

\subsection{Main Results and Analysis}
\label{sec:main_results}
\runin{Models set the tier.}
We group the 22 configurations of \cref{tab:main} into four tiers by their Overall score, with representative submissions shown in \cref{fig:submissions_grid}. Tier~1 holds all four configurations of GPT-6 Astra and Gemini 4 Argon, tier~2 includes both configurations of Claude Opus 5.5 and the Claude Code / Claude Sonnet 5.5 configuration, tier~3 the eleven others, and tier~4 all four configurations of GLM-5.3-Flash and Seed Doubao 2.1 Pro.

\runin{The harness effect varies by model.}
Changing a model's harness moves Overall by 0.023 on average in absolute value and by at most 0.056 (Claude Sonnet 5.5). Neither harness is better across the board: five models score higher under their vendor CLI and six under OpenCode, and the vendor CLI is 0.007 higher on average. A few models are sensitive to the harness: Claude Sonnet 5.5 falls from tier~2 to tier~3 under OpenCode, and its 0.056 and the 0.048 of GPT-6 Sol are comparable to the gains from the skills (\cref{sec:skills_eval}). Harnesses change $\bar{E}_1$--$\bar{E}_3$ about three times as much as $\bar{E}_4$.

\runin{Structure separates agents more than appearance.}
Across configurations, $\bar{E}_1$--$\bar{E}_3$ show score ranges of 0.40--0.43, compared with only 0.24 for $\bar{E}_4$. Geometry remains challenging, with tier~1 scoring 0.44--0.54 on $\bar{E}_3$. Agents differ less in how their scenes look than in where things are and what shape they have.

\runin{The dense-view task brings limited gains.}
The dense-view task scores only 0.009 higher on average, mainly in $E_1$--$E_3$, and both tasks rank the configurations almost identically. Since T2 differs from T1 in more than the number of views (\cref{sec:tasks}), this means only that agents gain little from the survey video under the current interface, in which they extract frames with \texttt{ffmpeg}, and that they use few of its frames (\cref{sec:skills_eval}).

\subsection{Efficiency Analysis}
\label{sec:efficiency}
\runin{Quality is weakly correlated with resource use.}
As \cref{fig:efficiency} shows, Overall is nearly unrelated to per-run tokens and time. The frontier runs from Claude Code / Claude Sonnet 5.5, the configuration with the fewest tokens and the shortest time per run, to Codex CLI / GPT-6 Astra, the highest-scoring one. A harness changes cost more than score: with the same model, the median run under the vendor CLI uses 0.49 to 1.87 times the tokens of the median run under OpenCode, while Overall changes by at most 0.056.

\runin{When a usable scene appears.}
Over the runs of \cref{fig:trajectory_models}, the first submission already reaches about 90\% of the final score on average, but it comes after 4.5 minutes for GPT-6 Sol and after more than half an hour for Gemini 4 Argon and HY-4, so ranking configurations by the score in hand during the hour can differ from ranking them by the final score. Some agents stop early: Claude Sonnet 5.5 within a few minutes, and GPT-6 Sol and DeepSeek-V4.1-Flash after about half an hour.

\begin{figure}[t]
\centering
\begin{minipage}[b]{0.50273\linewidth}
\centering
\includegraphics[width=\hsize]{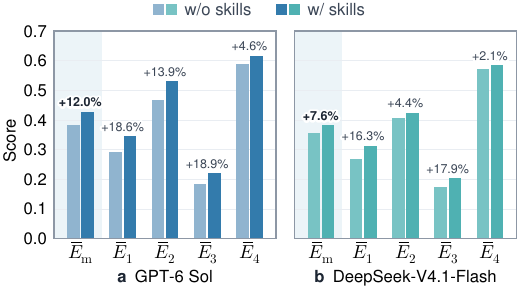}
\end{minipage}\hfill
\begin{minipage}[b]{0.47727\linewidth}
\centering
\includegraphics[width=\hsize]{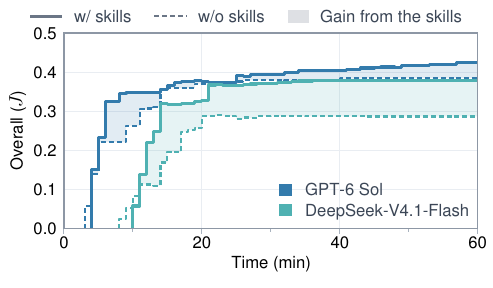}
\end{minipage}
\par\smallskip
\begin{minipage}[t]{0.50273\linewidth}
\caption{\textbf{Effect of the skills.} Mean scores of (a)~GPT-6 Sol and (b)~DeepSeek-V4.1-Flash under OpenCode on all 100 tasks, without (the runs of \cref{tab:main}) and with the three skills of \cref{sec:harness}. $\bar{E}_{\mathrm{m}}$ is the Overall score $J$ (\cref{eq:aggregate}); the numbers above the bars give the relative gain from the skills.}
\label{fig:skill_bars}
\end{minipage}\hfill
\begin{minipage}[t]{0.47727\linewidth}
\caption{\textbf{Score in hand with and without the skills.} Overall score $J$ of the latest submission at each minute, for GPT-6 Sol and DeepSeek-V4.1-Flash under OpenCode, without and with the three skills; each curve is the mean of 10 runs (as in \cref{fig:trajectory_models}).}
\label{fig:skill_trajectory}
\end{minipage}
\end{figure}

\subsection{Trajectory Analysis and Skill Transfer}
\label{sec:skills_eval}

\noindent\begin{minipage}[t]{0.575\linewidth}
\vspace{0pt}\setlength{\parskip}{6pt plus 2pt}%
\runin{What separates the tiers.}
As \cref{tab:behaviors} shows, tier~1 agents inspect the inputs more fully than tier~4 agents, compare their scenes with the inputs more often, and measure their scenes more accurately. They are also less likely to edit materials while a main item is still misplaced. Neither tier makes much use of the survey video. The three skills of \cref{sec:harness} target these practices; we next test whether they raise the score.

\runin{The skills raise the score.}
We give the three skills to two tier~3 models, GPT-6 Sol and DeepSeek-V4.1-Flash, under OpenCode, and run each task once. Overall rises from 0.383 to 0.429 (+12.0\%) for GPT-6 Sol and from 0.355 to 0.382 (+7.6\%) for DeepSeek-V4.1-Flash, and all four scores rise on average, with the largest relative gains in $\bar{E}_1$ and $\bar{E}_3$ (\cref{fig:skill_bars}); with the skills, DeepSeek-V4.1-Flash nearly matches GPT-6 Sol without them. Both models also reach high scores earlier (\cref{fig:skill_trajectory}). These results suggest that some tier~1 practices carry over to tier~3 models through the harness, without changing the models.
\end{minipage}\hfill
\begin{minipage}[t]{0.395\linewidth}
\vspace{0pt}%
\captionsetup{hypcap=false}%
\captionof{table}{\textbf{Behaviors behind the skills.} Means over all runs of tier~1 and tier~4. Premature material edits: share of material edits made while a main item is misplaced; scale error: deviation of the scene size from the ground truth.}
\label{tab:behaviors}
\footnotesize
\setlength{\tabcolsep}{5pt}
\centering
\begin{tabular}{@{}l rr@{}}
\toprule
Behavior & Tier 1 & Tier 4 \\
\midrule
\multicolumn{3}{@{}l}{\textbf{A\enspace Input processing}} \\
\quad Views opened (of 9) & 7.9 & 5.3 \\
\quad Pixel analysis (\% runs) & 94 & 59 \\
\quad Video frames used (\%) & 0.35 & 0.30 \\
\addlinespace[2pt]
\multicolumn{3}{@{}l}{\textbf{B\enspace Plan and update}} \\
\quad Reference-view checks & 17.9 & 13.9 \\
\quad Premature material edits (\%) & 32 & 81 \\
\addlinespace[2pt]
\multicolumn{3}{@{}l}{\textbf{C\enspace Photogrammetry and localization}} \\
\quad Building offset (m) & 1.7 & 4.0 \\
\quad Scale error (\%) & 5 & 38 \\
\quad Buildings located (\%) & 81 & 58 \\
\bottomrule
\end{tabular}\par

\end{minipage}

\section{Conclusion}
\label{sec:conclusion}

We presented SceneWright, a benchmark of multi-view procedural 3D modeling on 50 outdoor scenes that compares each submitted scene with its Blender ground truth in the cameras' frame, without registration. Across 22 configurations, the model sets the tier, the better harness depends on the model, and per-run tokens and time do not predict quality. Agents differ more in layout and geometry than in appearance, and under the current interface they gain little from a dense survey video over sparse views. The practices that separate the strongest agents partly transfer to weaker models as skills in the harness. The main gap for coding agents in 3D modeling is thus metric reasoning from calibrated views rather than visual plausibility, and SceneWright's registration-free scores measure progress on it for both models and harnesses. SceneWright can be extended in several directions. The scores can cover surfaces hidden from every view, such as interiors. The scenes can be complemented by real photographs and a wider range of natural scenes, and the survey video can be given to the model directly. On the agent side, the skills only begin to explore what a harness can contribute.

\end{document}